%% file: main.tex
\documentclass{article}

\input{source/preamble}
\usepackage[accepted]{source/icml2021}
\usepackage{makecell}

\icmltitlerunning{EE-559 Deep Learning: Group Mini-Project, Group 11}

\begin{document}

\twocolumn[
\icmltitle{From BERT to Qwen: Hate Detection across architectures}

\begin{icmlauthorlist}
\icmlauthor{Ariadna Mon Gomis}{ch} \quad
\icmlauthor{Saúl Fenollosa Arguedas}{ch} \quad
\icmlauthor{Jon Lecumberri Arteta}{ch} \\
\end{icmlauthorlist}

\icmlaffiliation{ch}{}
\vskip 0.05in
\centering
Group 11
\vskip 0.3in
]

\printAffiliationsAndNotice{} 

\begin{abstract}

Online platforms struggle to curb hate speech without over-censoring legitimate discourse. Early bidirectional transformer encoders made big strides, but the arrival of ultra-large autoregressive LLMs promises deeper context-awareness. Whether this extra scale actually improves practical hate-speech detection on real-world text remains unverified. Our study puts this question to the test by benchmarking both model families, classic encoders and next-generation LLMs, on curated corpora of online interactions for hate-speech detection (Hate or No Hate). 


\end{abstract}
\section{Introduction}
\label{sec:intro}

Online platforms face a mounting challenge in identifying and moderating hate speech at the speed and volume of user-generated content. Traditional approaches rely heavily on encoder-only transformer models such as BERT \cite{devlin2018bert} which, when fine-tuned on labeled hate speech corpora, achieve strong classification performance  \cite{saleh2021detection}. However, each new domain or task requires expensive retraining and careful hyperparameter tuning, often involving computationally intensive grid or random search \cite{heyamit2024finetuning}, thus making rapid deployment across diverse communities difficult.

Meanwhile, the advent of massive decoder-only large language models (LLMs) has demonstrated remarkable zero- and few-shot learning capabilities on a wide array of NLP tasks without full fine-tuning \cite{brown2020language}. These LLMs promise greater adaptability and lower maintenance cost: instead of retraining the entire model, users can simply craft new prompts (zero-shot) or supply a handful of labelled examples in context (few-shot). This prompt-based flexibility enables rapid prototyping across domains, though its effectiveness for sensitive tasks like hate speech detection remains under-explored.

In this regard, this project compares decoder-only LLMs under zero-shot, few-shot, and fine-tuning against fine-tuned encoder-only transformers on Mody et al.’s balanced hate-speech corpus \cite{mody2022curated}. We hypothesize that encoder-only models will outperform decoder-only models in zero-shot settings \cite{devlin2018bert,saleh2021detection}, that few-shot prompting will substantially boost decoder performance \cite{brown2020language,luca2023challenges}, and that task-specific fine-tuning of decoder-only models may allow them to rival or surpass encoder baselines \cite{brown2020language,heyamit2024finetuning}.

\section{Related Work}
\label{sec:related_work}

Studies that employ the Curated Hate Speech Dataset of Mody et al.\ \cite{mody2022curated} converge on a clear picture.  
Encoder-only transformers set the performance bar: Ishaq et al.\ retrained BERT into HateSpeechBERT and reported 96 \% $F_{1}$ \cite{ishaq2022hatespeechbert}.  
Zhou and Kavuri fine-tuned BERT to 83.8 \% accuracy, then attached a vision module to recover performance on obfuscated text \cite{zhou2023now}.  
Within the MetaHate benchmark, Piot et al.\ confirmed BERT-base as the strongest classical baseline at 0.88 $F_{1}$ \cite{piot2024metahate}.  
A large-scale survey by Abusaqer et al.\ placed RoBERTa-base above 90 \% $F_{1}$, with CatBoost and SVM trailing only a few points behind \cite{abusaqer2025efficient}.  
Sharif et al.\ achieved 89 \% accuracy with a lightweight CNN + BiLSTM, showing that hybrids can approach transformer quality while requiring less compute \cite{sharif2024enhancing}.

Despite these gains, two questions remain open.  
First, prompt-based decoder-only language models such as Qwen or Gemma have not been systematically evaluated on this dataset in zero-, few-, and full-shot regimes.  
Second, a cost-aware comparison between compact encoder models and small decoder-only LLMs is still lacking.

The present work fills these gaps by testing DistilBERT and Twitter-RoBERTa against Gemma-3-1B and Qwen-0.5B on identical splits, measuring zero-shot, few-shot, and fully fine-tuned performance to clarify the trade-off between prompt flexibility and task-specific optimisation.

\section{Dataset}
\label{sec:dataset}

Mody et al.’s curated hate speech dataset (Version 1) comprises 451.709 English sentences, of which 371.452 (approximately 82\%) are labelled as hateful and 80.250 (18\%) as non-hateful, drawn from a variety of online sources (Kaggle, GitHub, etc.) and reflecting real-world linguistic phenomena such as emoticons, emojis, hashtags, slang, and contractions \cite{mody2022curated}.
.

\section{Method}
\label{sec:method}

\subsection{Data Pre-processing}
\label{sec:preprocess}

\begin{figure}[h]
    \centering
    \includegraphics[width=\linewidth]{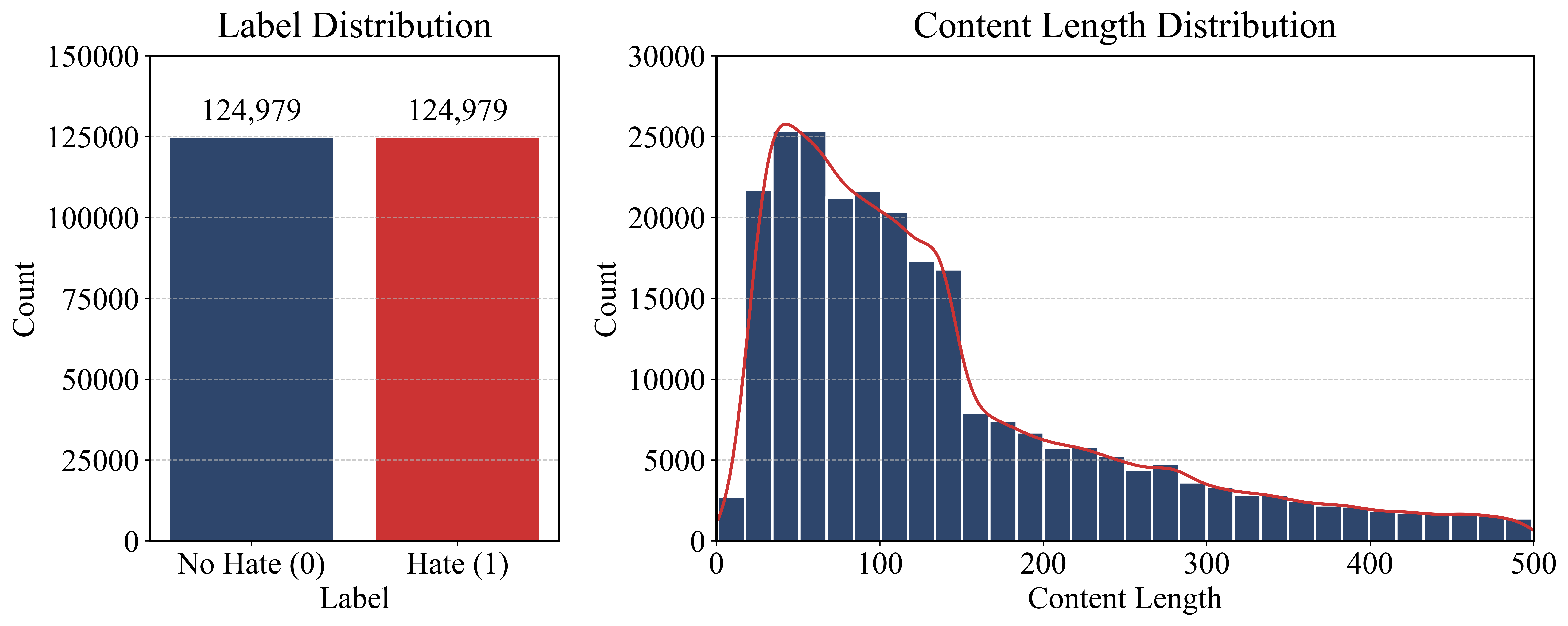}
    \vspace{-20pt}
    \caption{Label Distribution Balanced. Content Length Distribution.}
    \label{fig:eda}
\end{figure}

Before training, we first removed any entries exceeding 500 tokens to prevent extreme outliers from skewing training dynamics. Next, we addressed label imbalance by randomly undersampling the majority class and oversampling the minority class until both hateful and non-hateful classes were equally represented. After these operations, our final dataset consisted of over 100.000 high-quality, human-curated entries, balanced 50/50 across labels (see Fig. \ref{fig:eda}).

\subsection{Model Architectures}
\label{sec:models}

\begin{figure}[H]
    \centering
    \includegraphics[width=0.95\linewidth]{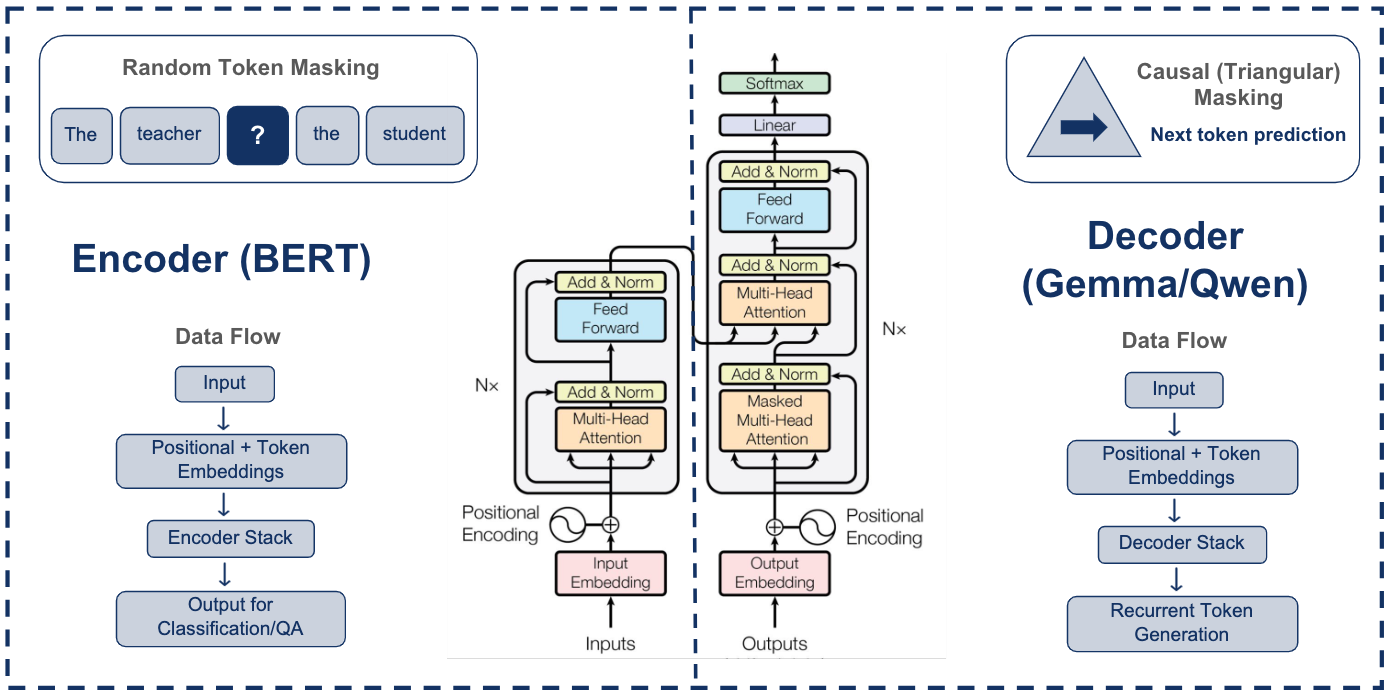}
    \caption{Transformers Architecture Comparison Overview.}
    \label{fig:arch}
\end{figure}

Figure \ref{fig:arch} highlights the key differences between our two model families. Encoder-only Transformers like BERT are trained with a masked-language‐model objective, allowing each token to attend to both its left and right context; this bidirectional design makes them well suited for fine-tuning on classification tasks. In contrast, decoder-only models use a causal attention mask, predicting one token at a time based solely on previous tokens, which enables autoregressive generation and allows classification via simple prompt-based queries without any gradient updates \cite{vaswani2017attention}.  

\subsubsection{Encoder‐Only Transformers}

We made use of two encoder‐only architectures, DistilBERT and Twitter-RoBERTa-base-offensive, using HuggingFace’s Transformers library \cite{distilbert-base-uncased,twitter-roberta-base-offensive}. DistilBERT is a distilled version of BERT-base that is 40\% smaller (66 M vs. 110 M parameters) and runs approximately 60\% faster, while retaining over 95\% of BERT’s performance \cite{xlnet2019generalized}. It has 6 transformer layers (vs. 12 in BERT-base), a hidden size of 768, and 12 attention heads. Alternatively, the Twitter-roBERTa-base-offensive used is a roBERTa-base model (12 layers, 768 hidden size, 12 heads, $\sim$125 M parameters), trained on $\sim$58M tweets and fine-tuned for offensive language identification with the TweetEval benchmark \cite{barbieri2020tweeteval}. These two encoder-only models were further fine-tuned with our dataset. 

\subsubsection{Decoder‐Only LLMs}

We also evaluate two lightweight decoder‐only models, Gemma-3 (1B) and Qwen1.5 (0.5B), via their Hugging Face model cards \cite{gemma_2025,qwen}. Gemma-3-1B is a 1 billion-parameter variant of Google’s Gemma 3 family, provided here in 4-bit quantized form.  Despite its small size, it inherits the same autoregressive Transformer architecture and supports up to 32K tokens of context, making it efficient for both text generation and few-shot classification tasks. Qwen-1.5-0.5B is a 500 million-parameter decoder-only model from the Qwen 1.5 series; this compact model is designed for fast inference and strong few-shot performance across languages \cite{qwen}. 

\subsection{Training Procedure}
\label{sec:training}

We used 70\% of the curated dataset for training, reserving the remainder for validation. Table \ref{tab:training} lists the transformer variants we evaluate, indicating their model name, whether they are fine-tuned, and their training hyperparameters.

\begin{table}[h]
    \vskip 0.15in
    \begin{center}
    \begin{small}
    \begin{sc}
    \begin{tabular}{c c c c}
    \toprule
    \textbf{Model} & \textbf{Fine‐tuned?} & \textbf{Train. Params} \\
    \midrule
     DistilBERT    & Yes  & \makecell[c]{2e, bs=128, \\ lr=\(2\times10^{-5}\)} \\
     roBERTa    & Yes     & \makecell[c]{2e, bs=128, \\ lr=\(2\times10^{-5}\)} \\
     Qwen1.5                     & Yes                  & \makecell[c]{3e, bs=64, \\ lr=\(2\times10^{-5}\)} \\
    Gemma‐3                     & No                   & –                    \\
    \bottomrule
    \end{tabular}
    \end{sc}
    \end{small}
    \end{center}
    \caption{Fine‐tuning and inference configurations for each Transformer model.}
    \label{tab:training}
    \vskip -0.1in
\end{table}

\section{Validation}
\label{sec:validation}

We set aside 30\% of the processed dataset as an untouched validation split and ran all evaluation experiments with five different random seeds (42, 123, 456, 789, 101112) to ensure statistical robustness; all resulting metrics have been averaged accordingly.

\begin{figure}[h]
  \centering
  \includegraphics[width=\linewidth]{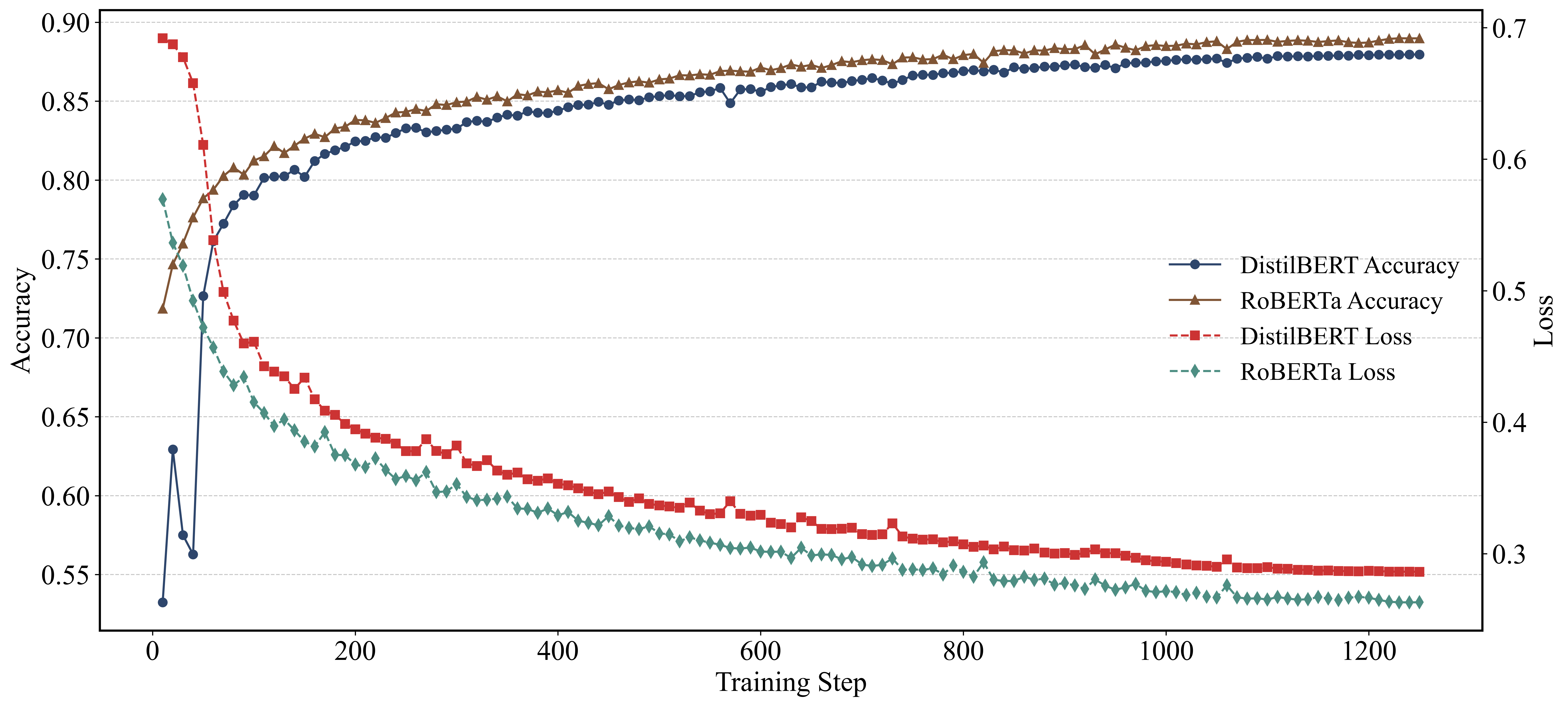}
  \vspace{-20pt}
  \caption{Validation accuracy and loss curves for DistilBERT vs.\ RoBERTa (seed 42).}
  \label{fig:acc_loss}
\end{figure}

For the encoder‐only models (DistilBERT, RoBERTa), we fine-tuned on the training split and monitored performance on the validation set. Figure \ref{fig:acc_loss} shows the accuracy and loss curves averaged across seeds. 

\begin{figure}[h]
  \centering
  \includegraphics[width=\linewidth]{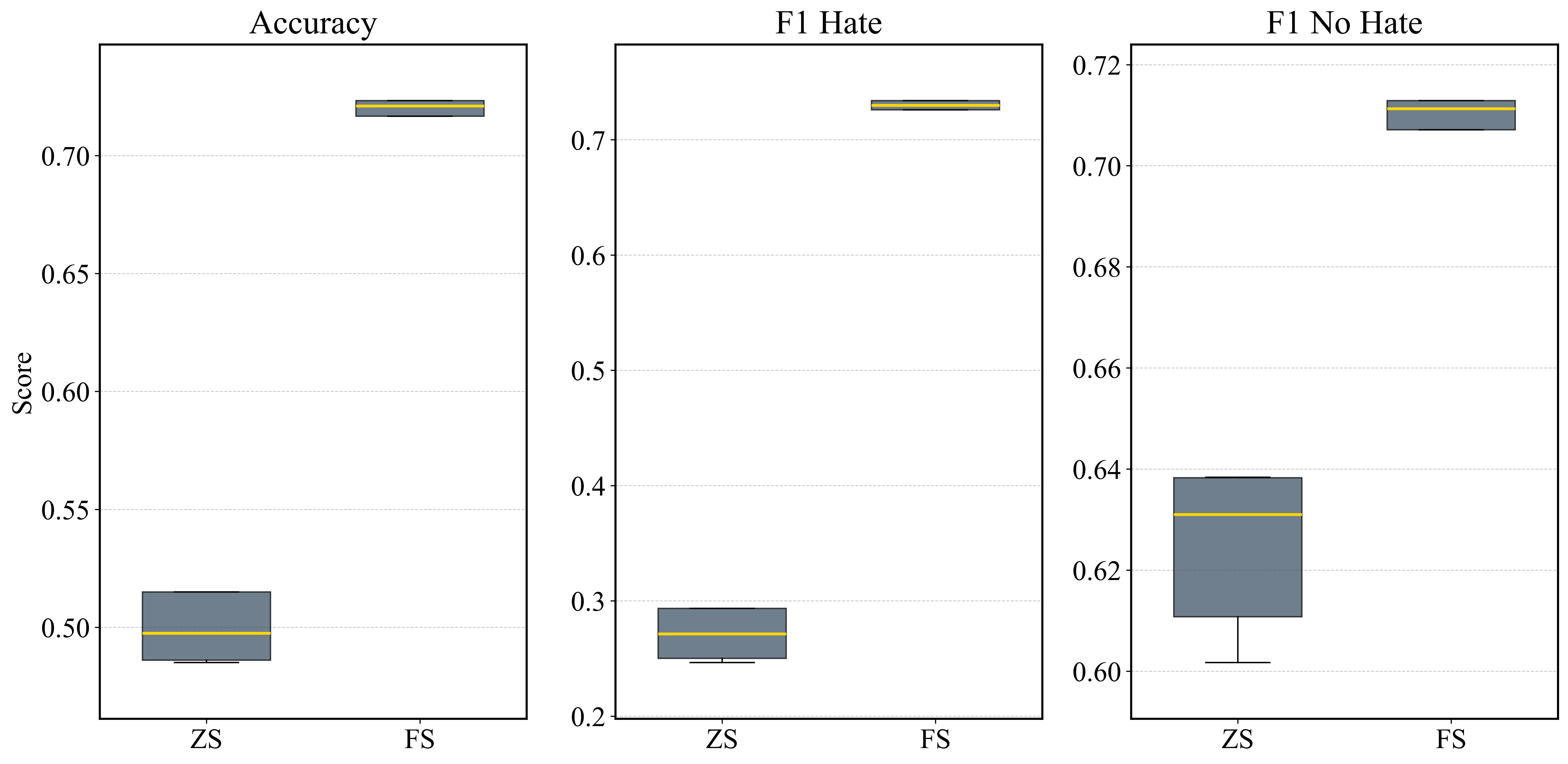}
  \vspace{-20pt}
  \caption{Base Gemma performance on the validation set under zero-shot (ZS) vs.\ few-shot (FS) prompting across three metrics (averaged over five seeds).}
  \label{fig:gemma_metrics}
\end{figure}

For decoder‐only models (Qwen, Gemma) we considered two evaluation settings on the same validation split: \textbf{zero-shot (ZS)}, no labelled examples provided in the prompt before inference, and \textbf{few-shot (FS):}, applied to Gemma only, where we prepend \(k=8\) dataset-related examples (4 Hate / 4 No-hate). We compare these settings in Figure \ref{fig:gemma_metrics}, which plots base Gemma’s ZS vs.\ FS performance across three standard metrics (Accuracy, F1-Hate, F1-No-Hate), aggregated over the five seeds.  

\begin{figure}[h]
  \centering
  \includegraphics[width=\linewidth]{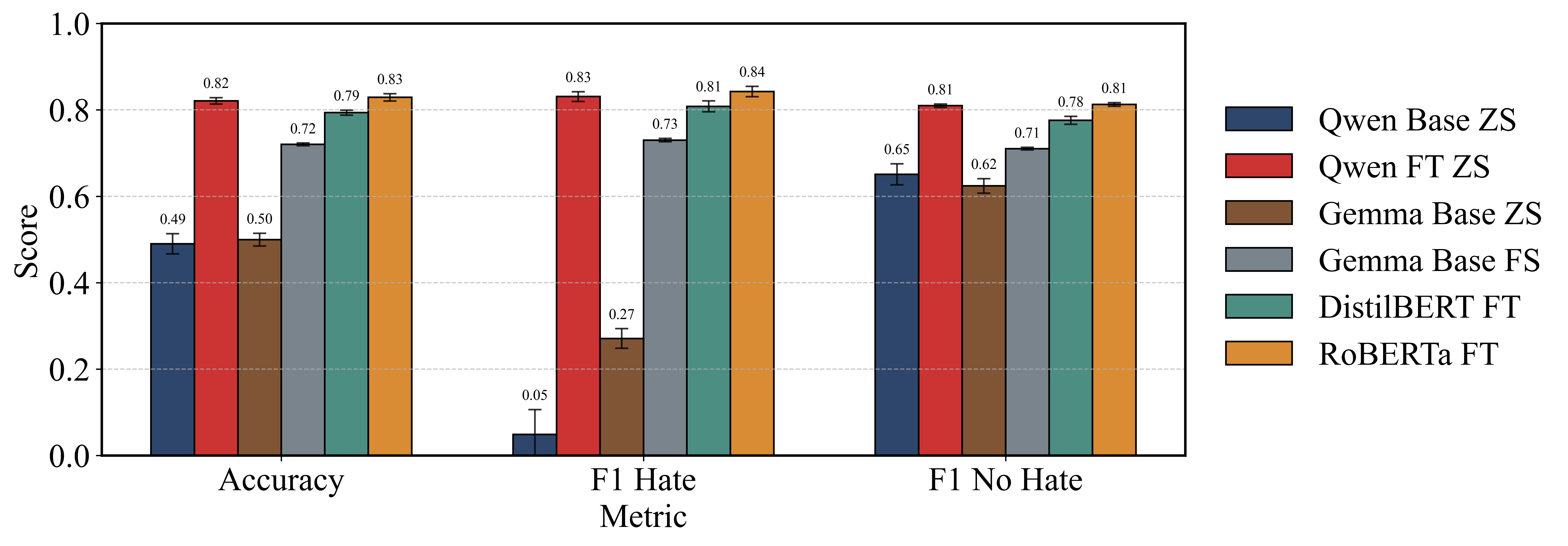}
  \vspace{-20pt}
  \caption{Final validation metrics for all models: Qwen base ZS, Qwen FT ZS, Gemma ZS, Gemma FS, DistilBERT FT, and RoBERTa FT.}
  \label{fig:all_metrics}
\end{figure}

Finally, Figure \ref{fig:all_metrics} summarizes the final validation metrics for all models, with error bars representing one standard deviation across seeds. 

The validation results (Figs. \ref{fig:acc_loss}, \ref{fig:all_metrics}) confirm several key insights; Qwen, when fine‐tuned and evaluated in ZS, delivers the highest overall accuracy and F1 scores, slightly edging out both encoder‐only baselines. This finding reinforces our hypothesis that decoder‐only models can match or exceed encoder‐only performance once they receive task‐specific weight updates. DistilBERT and RoBERTa follow closely behind, achieving F1 scores above 0.80 on both classes, while base Gemma performs poorly in ZS mode. However, when Gemma is supplied with FS, its performance jumps by over 0.15–0.20 F1 (Figure~\ref{fig:gemma_metrics}), supporting our hypothesis of few‐shot prompting potential. Our methodological choices ensure that these comparisons are fair and that observed differences stem from model architectures and inference regimes rather than data artifacts. The strong showing of Qwen FT ZS suggests that even relatively compact decoder‐only LLMs benefit substantially from discriminative fine‐tuning. Meanwhile, the FS gains for Gemma indicate that prompt‐based adaptation can partially compensate for the absence of gradient updates, though it remains sensitive to prompt design and example selection.

There are, however, important limitations. Our experiments are confined to a single English‐language corpus and have not been explored in other datasets, so the study might lack generalization. Moreover, computational times could not be compared consistently due to dependency issues and different runtime environments (Gemma/Qwen on separate VMs). Future work should explore simple hybrid approaches, mixing small, task-specific updates with prompt techniques, and test how well they hold up to adversarial examples and new data domains.

\section{Conclusion}
\label{sec:conclusion}

In this work, we systematically compared compact encoder‐only transformers (DistilBERT, RoBERTa) and lightweight decoder‐only LLMs (Qwen, Gemma) on a real‐world hate speech detection task, using zero‐shot, few‐shot, and fine‐tuned inference regimes under identical data splits and evaluation protocols. Our results show that fine‐tuned decoder‐only models can match or slightly exceed encoder‐only baselines, while few‐shot prompting substantially improves performance over pure zero‐shot settings. These findings highlight the versatility of decoder‐only architectures when combined with discriminative adaptation or minimal in‐context examples. Future extensions should investigate cross‐domain generalization, adversarial robustness, and cost‐effective hybrid methods to further bridge the gap between flexibility and accuracy.

\newpage


\bibliographystyle{ieeetr}
\bibliography{main}

\end{document}

%% file: source/preamble.tex
\usepackage{microtype}
\usepackage{graphicx}
\usepackage{subfigure}
\usepackage{booktabs} 

\usepackage{hyperref}

\usepackage[square,sort,comma,numbers]{natbib}